\documentclass{article}[11pt]  

\usepackage[utf8]{inputenc} 
\usepackage[T1]{fontenc}    
\usepackage{hyperref}       
\usepackage{url}            
\usepackage{booktabs}       
\usepackage{amsfonts}       
\usepackage{nicefrac}       
\usepackage{microtype}      

\usepackage{longtable,booktabs}
\usepackage{graphicx,float}

\title{Polynomial Regression as an Alternative to Neural Nets} 

\author{
  Xi Cheng \\
  Department of Computer Science\\
  University of California, Davis\\
  Davis, CA 95616, USA \\
  \texttt{xicheng0821@gmail.com}
  \and
  Bohdan Khomtchouk \\
  Department of Biology \\
  Stanford University \\
  Stanford, CA 94305, USA \\
  \texttt{bohdan@stanford.edu} \\
  \and
  Norman Matloff \\
  Department of Computer Science\\
  University of California, Davis \\
  Davis, CA 95616, USA \\
  \texttt{matloff@cs.ucdavis.edu} \\
  \and
  Pete Mohanty \\
  Department of Statistics \\
  Stanford University \\
  Stanford, CA 94305, USA \\
  \texttt{pmohanty@stanford.edu} 
} 

\begin{document}

\maketitle

\begin{abstract}

\noindent Despite the success of neural networks (NNs), there is still a
concern among many over their ``black box'' nature. Why do they work? 
Yes, we have Universal Approximation Theorems, but these concern
statistical consistency, a very weak property, not enough to explain the
exceptionally strong performance reports of the method.  Here we present
a simple analytic argument that NNs are in fact essentially polynomial
regression models (PR), with the effective degree of the polynomial
growing at each hidden layer.  This view will have various implications
for NNs, e.g.\ providing an explanation for why convergence problems
arise in NNs, and it gives rough guidance on avoiding overfitting.  In
addition, we use this phenomenon to predict and confirm a
multicollinearity property of NNs not previously reported in the
literature.  Most importantly, this NN $\leftrightarrow$ PR
correspondence suggests routinely using polynomial models instead of
NNs, thus avoiding some major problems of the latter, such as having to
set many hyperparameters and deal with convergence issues.  We present a
number of empirical results; in all cases, the accuracy of the
polynomial approach matches, and often exceeds, that of NN approaches.
A many-featured, open-source software package, {\bf polyreg}, is
available.


\end{abstract}

\section{The Mystery of NNs}

Neural networks (NNs), especially in the currently popular form of
many-layered {\it deep learning networks} (DNNs), have become many
analysts' go-to method for predictive analytics.  Indeed, in the popular
press, the term {\it artificial intelligence} has become virtually
synonymous with NNs.\footnote{In this paper, we use the term ``NNs'' to
mean general feedforward networks, as opposed to specialized models such
as Convolutional NNs.  See Section \ref{specialized}.}

Yet there is a feeling among many in the community that NNs are ``black
boxes''; just what is going on inside?  Various explanations have been
offered for the success of NNs, e.g.\ \cite{Shwartz}.
However, the present paper will report significant new insights.

\section{Contributions of This Paper}

The contribution of the present work will be as follows:\footnote{
Author listing is alphabetical by surname.  Individual contributions, in
roughly chronological order, have been: NM conceived of the main ideas
underlying the work, developed the informal mathematical
material, and wrote support code for \textbf{polyreg}; XC wrote the
core code for {\bf polyreg}; BK assembled the brain and kidney
cancer data, and provided domain expertise guidance for genetics
applications; PM rewrote the entire polynomial generator in
\textbf{polyreg}, extended his {\bf kerasformula} package for use with
it; and provided specialized expertise on NNs.  All authors conducted
data experiments.  The authors also appreciate some assistance by
Matthew Kotila and Allen Zhao.}  

\begin{itemize}  

   \item [(a)] We will show that, at each layer of an NN, there is a
   rough correspondence to some fitted ordinary parametric polynomial
   regression (PR) model; {\bf in essence, NNs are a form of PR}.  We
   refer to this loose correspondence here as NN $\leftrightarrow$ PR. 

   \item [(b)] A very important aspect of NN $\leftrightarrow$ PR is
   that {\bf the degree of the approximating  polynomial increases with
   each hidden layer}.  In other words, our findings should not be
   interpreted as merely saying that the end result of an NN can be
   approximated by some polynomial.

   \item [(c)] We {\bf exploit NN $\leftrightarrow$ PR to learn about
   general properties of NNs via our knowledge of the properties of PR.}
   This will turn out to provide new insights into aspects such as the
   number of hidden layers and numbers of units per layer, as well as
   how convergence problems arise.  For example, we use NN
   $\leftrightarrow$ PR to \textbf{predict and confirm a multicollinearity
   property of NNs not previous reported in the literature.}

   \item [(d)] Property (a) suggests that in many applications, one
   might simply fit a polynomial model in the first place, bypassing
   NNs.  This would have the advantage of {\bf avoiding the problems of
   choosing numerous hyperparameters}, nonconvergence or convergence to
   non-global minima, and so on.

   \item [(e)] Acting on point (d), we compare NN and polynomial models
   on a variety of datasets, finding {\bf in all cases that PR gave results
   at least as good as, and often better than, NNs}.\footnote{In view of
   the equivalence, one may ask why PR often outperforms NNs.  The
   answer is that }

   \item [(f)] Accordingly we have developed an open source,
   feature-rich software package in R (a Python version is planned),
   {\bf polyreg}, that enables active implementation of the above ideas
   \cite{xi}.\footnote{Developing an algorithm to generate the
   polynomials is a nontrivial task, as one must deal with the fact that
   many applications include a number of categorical variables.  Since
   these must be converted to dummy variables, i.e.\ vectors consisting
   of 1s and 0s, one should not compute powers of such variables, as
   the powers equal the original vector.  Similarly, since the product
   of two dummies arising from the same categorical variable is
   necessarily the 0 vector, we should not compute/store these products
   either.

   }

\end{itemize}  

Point (a) is especially important, as it shows a connection of NNs to PR
that is much tighter than previously reported.  Some researchers, e.g.\
\cite{ong}, have conducted empirical investigations of the possible use
of polynomial regression in lieu of NNs.  Some theoretical connections
between NNs and polynomials have been noted in the literature, e.g.\
\cite{hornik}.  Furthermore, some authors have constructed networks
consisting of AND/OR or OR/AND polynomials as alternatives to NNs
\cite{shin}.  \cite{benitez} showed an explicit equivalence of NNs to
fuzzy rule-based systems, and \cite{rational} derived a similar
correspondence between NNs and rational functions. 

But our contribution goes much deeper.  It shows that in essence,
\textbf{conventional NNs actually \underline{are} PR models}.  Our focus
will be on the activation function.  Using an informal mathematical
analysis on that function, we show {\it why} NNs are essentially a form
of PR.  

Moreover, we stress the implications of that finding.  Indeed, in
Section \ref{lurk}, we will use our knowledge of properties of
polynomial regression to predict and confirm a corresponding property of
NNs that, to our knowledge, has not been reported in previous
literature.

\section{What This Paper Is NOT}

\subsection{The Paper Is Not Another Universal Approximation Theorem} 

Any smooth regression/classification, say continuous and bounded, can be
approximated \textit{either} by NNs (Universal Approximation Theorem)
\cite{poggio1989theory} \cite{hornik}, \textit{or} by polynomials
(Stone-Weirerstrass Theorem) \cite{rudin}, and it may at first appear
that our work here simply reflects that fact.  But we show a subtler but
much stronger connection than that.  We are interested in the NN fitting
process itself; we show that \textbf{fitting NNs actually mimics PR,
with higher and higher-degree polynomials emerging from each successive
layer.}

\subsection{The Paper Is Not about Specialized NNs}
\label{specialized}

Our work so far has primarily been on general feedforward NNs.  Our
investigations have not yet involved much on specialized networks such
as convolutional NNs (CNNs) for image classification, recurrent NNs
(RNNs) for text processing, and so on.  Though we intend to adapt our
ideas to these frameworks, \textbf{we view them as separate, orthogonal
issues}.

For instance, we view the convolutional ``front ends'' in CNNs as
largely playing the role of preprocessing stages, conducted for
dimension reduction.  As such, they are easily adaptable to other
approaches, including polynomial models.  Indeed, convolutional versions
of random forests have been developed \cite{Zhou} \cite{Miller}.  As
\cite{Zhou} points out, a key to the success of CNNs has been a general
property that is not specific to neural-learning paradigms (emphasis
added):

\begin{quote}

...the mystery behind the success of deep neural networks owes much to
three characteristics, i.e., layer-by-layer processing, {\bf in-model feature
transformation} and sufficient model complexity.

\end{quote}

Similar points hold for RNNs.  For instance, one might use {\it
structural equation models} \cite{monecke} with polynomial forms.  

We do include one brief, simple image classification example using the
MNIST dataset.\footnote{This is in Section \ref{mnist}.  It should not be
confused with the example in Section \ref{letters}, which involves image
classification with the data \textit{already} preprocessed, or with two
examples using MNIST to illustrate other phenomena.} Though
certainly a topic for future work, we view the preprocessing issue as
separate from our findings that NNs are essentially PR models, with
various important implications, and that PR models perform as well as,
if not better than, NNs.  

\subsection{The Paper Is Not about Other ML Methods}  

It may well be that PR also has relations to random forests, support
vector machines and so on.  However, our purpose here is not to seek
possible relations of PR to other methods, nor is it our purpose to
compare the performance of those other methods to PR and NNs.
Similarly, though we find for the datasets investigated here, PR does as
well as or better than NNs, there is no implied claim that PR would
perform well in comparison to other ML methods.

Instead, we note that NN $\leftrightarrow$ PR implies a very direct
relationship between PR and NNs, and explore the consequences.

\section{Notation}

Consider prediction of \(Y\) from a vector \(X\) of \(p\) features.  In
regression applications, \(Y\) is a continuous scalar, while in the
classification case it is an indicator vector, with \(Y_i = 1, Y_j = 0
\textrm{ for } j \neq i\) signifying class \(i\).  Prediction requires
first estimating from data the regression function \(r(t) = E(Y
| X = t)\), which in the classification case is the vector of
conditional class probabilities.  The function $r(t)$ must be estimated,
parametrically or nonparametrically, from sample data consisting of $n$
cases/observations.

\section{Polynomial Regression Models}

PRs --- models that are linear in parameters but polynomial in the
predictors/features --- are of course as old as the linear model itself.
(And they extend naturally to generalized linear models.) Though they
might be introduced in coursework for the case $p = 1$, multivariate
polynomial models are popular in {\it response surface methods}
\cite{montgomery}.

One issue of concern is {\it multicollinearity}, correlations among the
predictors/features \cite{faraway}.  PR models are long known to suffer
from multicollinearity at high degrees \cite{chatterjee}.

Indeed, in the case $p = 1$, fitting a polynomial of degree $n-1$ will
be an ephemeral ``perfect fit,'' with $R^2 = 1$.  Then any variable,
predictor or response, will be an exact linear combination of the
others, i.e.\ full multicollinearity.

For this and other reasons, e.g.\ large fitted values at the edges of
the data, many authors recommend not using polynomial models of degree
higher than 2 or 3, and in fact in our empirical experiments in this
paper, we have usually not needed to use a higher degree.  

On the other hand, it is important to note that this is not a drawback
of PR relative NNs.  On the contrary, due to NN $\leftrightarrow$ PR,
NNs have the same problem on the edges of the data.

\section{The NN $\leftrightarrow$ PR Principle}
\label{why} 

Universal Approximation Theorems (UATs) such as \cite{hornik} show that,
under various regularity assumptions and sufficient data, NNs can
approximate the regression function \(r(t)\) to any desired degree of
accuracy.  But this is essentially just a \textit{statistical
consistency} property; many statistical estimators are consistent but
perform poorly.  For example, consider the simple problem of estimating
the mean $\mu$ of a Gaussian distribution with known variance.  The
sample median is a statistically consistent estimator of $\mu$, but it
is only $2/\pi$ as \textit{efficient} as the sample mean
\cite{dasgupta}.\footnote{The sample mean achieves the same statistical
accuracy using only $2/\pi$ as much data.}  (See also Section
\ref{crossfit}.) So, while of theoretical interest, UATs are not too
useful in practice.

Instead, we aim for a much stronger property.

\subsection{The Basic Argument}

Take the simple case \(p = 2\).  Denote the features by \(u\) and \(v\).
The inputs to the first hidden layer, including from the ``1'' node,
will then be of the form $a_{00} + a_{01}u +a_{02}v$ and $a_{03} +
a_{05}u +a_{05}v$.

As a toy example, take the activation function to be \(a(t) = t^2\).
Then outputs of that first layer will be quadratic functions of $u$ and
$v$.  Similarly, the second layer will produce fourth-degree
polynomials, then degree eight, and so on.  Clearly, this works the same
way for any polynomial activation function.  So, for a polynomial
activation function, minimizing an $L_2$ loss function in fitting an NN
is the same as PR.

Now let's turn to more realistic activation functions, say
transcendental functions such as $\tanh$.  Computer implementations
often use Taylor series.   This was recommended in \cite{nyland} for
general usage and was used in firmware NN computation in
\cite{temurtas}.  So in such cases, the situation reverts to the above,
and NNs are still exactly PR.

In these cases, we have:

\begin{quote}

If the activation function is any polynomial, or is implemented by one,
an NN exactly performs polynomial regression.

Moreover, the degree of the polynomial will increase from layer to
layer.  

\end{quote}

For general activation functions and implementations, we can at least
say that the function is close to a polynomial, by appealing to 
the famous Stone-Weierstrass Theorem \cite{rudin}, which states that any
continuous function on a compact set can be approximated uniformly by
polynomials.\footnote{Compact sets are bounded, and indeed almost any
application has bounded data. No human is 200 years old, for instance.}
In any event, for practical purposes here, we see that most activation
functions can be approximated by a polynomial.  Then apply the same
argument as above, which implies: 

\begin{itemize}  

\item [(a)] {\bf NNs can loosely be viewed as a form of polynomial
regression}, our NN $\leftrightarrow$ PR Principle introduced earlier.  

\item [(b)] The degree of the approximating polynomial {\bf increases from
layer to layer}.

\end{itemize}

\subsection{What about ReLU?}

One of the most popular choices of activation function is ReLU,
$a(t) = \max(0,t)$.  It does not have a Taylor series, but as above, it
can be closely approximated by one.  Better yet, it is a piecewise
polynomial.  That implies that as we go from layer to layer, NN is
calculating a piecewise polynomial of higher and higher degree.  In
other words, we could write NN $\leftrightarrow$ PPR, the latter
standing for piecewise polynomial (or even piecewise linear) regression. 

\subsection{Some Elaboration}

The informal arguments above could be made mathematically rigorous.  In
that manner, we can generate polynomials which are dense in the space of
regression functions.  But let's take a closer look.

Suppose again for the moment that we use a polynomial activation
function.  The above analysis shows that, in order to achieve even
statistical consistency, the number of hidden layers must go to infinity
(at some rate) as $n$ grows; otherwise the degree of the fitted
polynomial can never go higher than a certain
level.\footnote{Specifically the degree of the activation function times
the number of layers.} But for a general activation function, a single
layer suffices, providing the number of neurons goes to infinity.

%
%

\section{Lurking Multicollinearity}
\label{lurk}


As noted, PR models of course have a very long history and are
well-understood.  We can leverage this understanding and the NN
$\leftrightarrow$ PR Principle to learn some general properties of NNs.
In this section, we will present an example, with intriguing
implications. 

As mentioned, PR models tend to produce multicollinearity at higher
degrees.  The material on NN $\leftrightarrow$ PR in the preceding
section, viewing NNs as kind of a polynomial regression method, suggests
that NNs suffer from multicollinearity problems as well. 


Indeed, the conceptual model of the preceding section would suggest that
the outputs of each layer in an NN become more collinear as one moves
from layer to layer, since multicollinearity tends to grow with the
degree of a polynomial.  

One then needs a measure of multicollinearity.  A very common one is the
{\it variance inflation factor} (VIF) \cite{faraway}.\footnote{Various
other measures of multicollinearity have been proposed, such as {\it
generalized variance}.} When running a linear regression analysis
(linear in the coefficients, though here polynomial in the
predictors/features), a VIF value is computed for each coefficient.
Larger values indicate worse degrees of multicollinearity.  There are no
firm rules for this, but a cutoff value of 10 is often cited as cause
for concern.

\subsection{Experimental Results}

For convenience, we used the MNIST data, with the {\bf keras} package.
Our test consisted of sequential models containing linear stacks of
layers, with five layers in total.  This included two dropout layers.
We set the number of units to be equal in each layer.  We calculated two
measures of overall multicollinearity in a given NN layer: the
proportion of coefficients with VIF larger than 10 and the average VIF. 


{\it First experiment:}
The number of units is 10 in each layer, and the model is the following. 

\begin{verbatim}
Layer (type)                    Output Shape                  Param #        
============================================================================
dense_1 (Dense)                (None, 10)                      7850           
____________________________________________________________________________
dropout_1 (Dropout)             (None, 10)                      0              
____________________________________________________________________________
dense_2 (Dense)                (None, 10)                      110            
____________________________________________________________________________
dropout_2 (Dropout)            (None, 10)                      0              
____________________________________________________________________________
dense_3 (Dense)                (None, 10)                      110            
============================================================================
\end{verbatim}

The VIF results are shown in Table \ref{model1}. On average, VIF
increases as one moves on to the next layer. 

{\it Second experiment:} We set the number of units to 64 in the first
four layers, while the last layer still has 10 outputs. The model is the
following. 

\begin{verbatim}
Layer (type)                    Output Shape                  Param #        
============================================================================
dense_1 (Dense)                (None, 64)                      50240           
____________________________________________________________________________
dropout_1 (Dropout)             (None, 64)                      0              
____________________________________________________________________________
dense_2 (Dense)                (None, 64)                      4160            
____________________________________________________________________________
dropout_2 (Dropout)            (None, 64)                      0              
____________________________________________________________________________
dense_3 (Dense)                (None, 10)                      650            
============================================================================
\end{verbatim}

The results of the VIF values of coefficients are shown in Table \ref{model2}. 

{\it Third experiment:} We set the number of units to 128 in the first
four layers and the last layer still has 10 outputs. The model is the
following. 

\begin{verbatim}
Layer (type)                    Output Shape                  Param #        
============================================================================
dense_1 (Dense)                (None, 128)                      100480           
____________________________________________________________________________
dropout_1 (Dropout)             (None, 128)                      0              
____________________________________________________________________________
dense_2 (Dense)                (None, 128)                      16512            
____________________________________________________________________________
dropout_2 (Dropout)            (None, 128)                      0              
____________________________________________________________________________
dense_3 (Dense)                (None, 10)                      1290            
============================================================================
\end{verbatim}

The results of the VIF values of coefficients are shown in Table \ref{model3}.  

\begin{table}
	\caption{Results of first model}
	\label{model1}
	\centering
	\begin{tabular}{c c c}
		Layer     & Percentage of VIFs that are larger than 10 & Average VIF \\
		\midrule
		dense\_1     & 0  &  3.43303    \\
		dropout\_1  & 0  & 3.43303      \\
		dense\_2     & 0.7  & 14.96195  \\
		dropout\_2  & 0.7  & 14.96195 \\
		dense\_3     & 1   & $1.578449 \times 10^{13}$ \\
		\bottomrule
	\end{tabular}
\end{table}

\begin{table}
	\caption{Results of second model}
	\label{model2}
	\centering
	\begin{tabular}{c c c} 
		Layer     & Percentage of VIFs that are larger than 10 & Average VIF \\
		\midrule
		dense\_1     & 0.015625  &  4.360191    \\
		dropout\_1  & 0.015625  & 4.360191      \\
		dense\_2     & 0.96875  & 54.39576  \\
		dropout\_2  & 0.96875  & 54.39576 \\
		dense\_3     & 1   & $3.316384 \times 10^{13}$ \\
		\bottomrule
	\end{tabular}
\end{table}

\begin{table}
	\caption{Results of third model}
	\label{model3}
	\centering
	\begin{tabular}{c c c}
		Layer     & Percentage of VIFs that are larger than 10 & Average VIF \\
		\midrule
		dense\_1     & 0.0078125  &  4.3537    \\
		dropout\_1  & 0.0078125  & 4.3537      \\
		dense\_2     & 0.9921875  & 46.84217  \\
		dropout\_2  & 0.9921875  & 46.84217 \\
		dense\_3     & 1   & $5.196113 \times 10^{13}$ \\
		\bottomrule
	\end{tabular}
\end{table}

\subsection{Impact}

In the above experiments, the magnitude of multicollinearity increased
from layer to layer. This increasing multicollinearity corresponds to
the multicollinearity warning in polynomial regression. Thus, NNs and
polynomial regression appear to have the same pathology, as expected
under NN $\leftrightarrow$ PR.   

In other words, {\bf NNs can suffer from a hidden multicollinearity
problem}.  This in turn is likely to result in NN computation
convergence problems.

We thus believe it would be helpful for NN software to include layer-by-layer
checks for multicollinearity.  If a layer is found to output a higher
degree of multicollinearity, one might consider reducing the number of
units in it, or even eliminating it entirely.  Applying dropout to such
layers is another possible action.  One related implication is that
later NN layers possibly should have fewer units than the earlier ones.  

It also suggests a rationale for using {\it regularization} in NN
contexts, i.e.\ shrinking estimators toward 0 \cite{hastie, nmregbook}.
The first widely-used shrinkage estimator for regression, {\it ridge
regression}, was motivated by amelioration of multicollinearity.  The
above discovery of multicollinearity in NNs provides at least a partial
explanation for the success of regularization in many NN applications.
Again due to NN $\leftrightarrow$ PR, this is true for PR models as
well.  We intend to add a ridge regression option to {\bf polyreg}.

Much more empirical work is needed to explore these issues.  

\section{PR as Effective, Convenient Alternative to NNs}

We compared PR to NNs on a variety of datasets, both in regression and
classification contexts (i.e. continuous and categorical response
variables, respectively).  The results presented here are complete,
representing every analysis conducted by the authors, i.e.\ not just the
``good'' ones.\footnote{We also started an analysis of the Missed
Appointments Data on Kaggle,
https://www.kaggle.com/joniarroba/noshowappointments.  However, we
abandoned it because no model improved in simply guessing No
(appointment not missed).  However, PR and NNs did equally well.}
However, not all hyperparameter combinations that were run are
presented; only a few typical settings are shown. Generally the settings
that produced extremely poor results for NNs are not displayed.

Each table displays the results of a number of settings, with the latter
term meaning a given method with a given set of hyperparameters.
For each setting: 

\begin{itemize}  

\item The dataset was split into training and test sets, with the number
of cases for the latter being the min(10000,number of rows in full set).

\item The reported result is mean absolute prediction error (MAPE) in
the regression case and overall proportion of correct classification
(PCC) in the classification case.

\end{itemize}  

No attempt was made to clean the data, other than data errors that
prevented running the code.  Note that this is especially an issue with
the NYC taxi dataset.  Performance may be affected accordingly.

The reader will recognize a number of famous datasets here, many from
the UC Irvine Machine Learning Repository.  There are also some ``new''
datasets, including: a specialized Census dataset on Silicon Valley
programmer and engineer wages, curated by one of the authors; data on
enrollments in Massive Open Online Courses (MOOCs); data from a Crossfit
competition; and data exploring the impact of genetics on brain and
kidney cancers, curated by another of the authors.\footnote{Note: Since
the time these experiments were done, \textbf{polyreg} has been
revamped.  The numbers here may change somewhat when we re-run the
experiments under the new version.}

Abbreviations in the tables:

\begin{itemize}  

\item PR:  Polynomial regression.  Degree is given, and if not the same,
maximum interaction term degree.  A ``PCA'' designation means that
dimension reduction via 90\%-total-variance Principal Components 
Analysis was performed before generating the polynomials.

\item FSR: Forward Stepwise Regression (part of {\bf polyreg}). Implementation
of PR which adds features and interactions (including polynomial terms, 
which can be thought of as self-interactions) one at a time, easy memory constraints. 

\item KF:  NNs through the Keras API, \cite{chollet2015keras}, accessed
in turn via the R-language package {\bf kerasformula} \cite{pete}.  The
default configuration is two layers with 256 and 128 units (written as
``layers 256,128''), and dropout proportions of 0.4 and 0.3.  In our
experiments, we varied the number of units and dropout rate, and used
ReLU and either 'softmax' or 'linear' for the prediction layer, for
classification and regression, respectively.

\item DN:  NNs through the R-language package {\bf deepnet}
\cite{xiaorong}.  The notation is similar to that of KF.  We used the
defaults, except that we took {\bf output} to be 'linear' for regression
and 'softmax' for classification.

\end{itemize}  

DN can be much faster (if less flexible) than KF, and thus DN was
sometimes used in the larger problems, or for comparison to
KF.\footnote{Concerning speed, KF does have a GPU version available; DN
does not.} However, their performance was similar.

The best-forming platform is highlighted in bold face in each case.

\subsection{Programmers and Engineers Census Data}

This is data on programmers and engineers in Silicon Valley in the 2000
Census.  There are 20090 rows and 16 columns.

First, we predict wage income, a regression context.  The results are
shown in Table \ref{prgengwg}.
We then predict occupation (six classes), shown in Table
\ref{prgengocc}.  Here PR substantially outperformed NNs. 

\begin{table}
\caption{Prg/Eng, predict income}
\begin{center}
\begin{tabular}{|l|r|}
\hline
setting & MAPE \\ \hline 
PR, 1 & 25595.63 \\ \hline
PR, 2 & 24930.71 \\ \hline
PR, 3,2 & 24586.75 \\ \hline
PR, 4,2 & {\bf 24570.04} \\ \hline
KF, default & 27691.56 \\ \hline
KF, layers 5,5  & 26804.68   \\ \hline
KF, layers 2,2,2  & 27394.35   \\ \hline
KF, layers 12,12  & 27744.56   \\ 
\hline
\end{tabular}
\end{center}
\label{prgengwg}
\end{table}

\begin{table}
\caption{Prg/Eng, predict occ.}
\begin{center}
\begin{tabular}{|l|r|}
\hline
setting & PCC \\ \hline 
PR, 1 & 0.3741 \\ \hline 
PR, 2 & {\bf 0.3845} \\ \hline 
KF, \textrm{default} & 0.3378 \\ \hline
KF, layers 5,5 & 0.3398 \\ \hline
KF, layers 5,5; dropout 0.1 & 0.3399 \\ \hline
\end{tabular}
\end{center}
\label{prgengocc}
\end{table}

\subsection{Million Song Data}

This is a very well-known dataset, listing audio characteristics of
songs, along with their year of publication.  The latter is the object
of prediction.  In this version of the dataset, there are 515345 cases,
with 90 predictor variables.  The results are shown in Table
\ref{millsong}.  PR was somewhat ahead of NNs in this case.

Note that the PR experiments used PCA.  A current limitation of PR in
{\bf polyreg} is that memory/time can become an issue, which occurred
here.  

\begin{table}
\caption{Million Song, predict year}
\begin{center}
\begin{tabular}{|l|r|}
\hline
setting & MAPE \\ \hline 
PR, 1, PCA & 7.7700\\ \hline 
PR, 2, PCA & {\bf 7.5758} \\ \hline 
KF, default & 8.4300 \\ \hline
KF, units 5,5; dropout 0.1,0.1 & 7.9883\\ \hline
KF, units 25,25; dropout 0.1,0.1 & 7.9634\\ \hline
KF, units 100,100; dropout 0.1,0.2 & 8.1886\\ \hline
KF, units 50,50,50,50; dropout 0.1,0.1,0.1,0.2 & 8.0129\\ \hline
KF, units 50,50,50,50,50; dropout 0.1,0.1,0.1,0.1,0.2 & 8.0956\\ \hline
KF, units 10,10,10,10,10,10; dropout 0.1,0.1,0.1,0.1,0.2 & 8.1102\\ \hline
DN, layers 5,5 & 8.7043 \\ \hline
DN, layers 8,2 & 9.5418 \\ \hline
DN, layers 2,2 & 7.8809 \\ \hline
DN, layers 3,2 & 7.9458 \\ \hline
DN, layers 3,3 & 7.8060 \\ \hline
DN, layers 2,2,2 & 8.7796 \\ \hline
\end{tabular}
\end{center}
\label{millsong}
\end{table}

\subsection{Concrete Strength Data}

Here one is predicting compressive strength of concrete.  This dataset
provides some variety in our collection, in that it is much smaller,
only 1030 rows.  There are eight predictors.

In Table \ref{conctable}, we see that PR significantly outperformed both
{\bf kerasformula} and the {\bf neuralnet} package.  This is probably to
be expected in a small dataset.  (Mean absolute error is not reported in
this case; the displayed values are correlations between predicted and
actual values, the square root of $R^2$.)  

\begin{table}
\caption{Concrete, predict strength}
\begin{center}
\begin{tabular}{|l|r|}
\hline
method & correlation (pred. vs. actual) \\ \hline
neuralnet & 0.608 \\ \hline
kerasformula & 0.546 \\ \hline
PR, 2 & {\bf{0.869}} \\ 
\hline
\end{tabular}
\end{center}
\label{conctable}
\end{table}

\subsection{Letter Recognition Data}
\label{letters}

As noted, we view preprocessing in image classification applications to
be orthogonal to the issues we are discussing.  But there is another
UCI dataset, which is already preprocessed, and thus was easy to include
in our experiments here.

The data consist of images of letters, preprocessed to record 16
geometric features.  There are 20000 images.  In spite of our attempts
with various combinations of hyperparameters, the performance of NNs,
both KF and DN, here was poor, and PR was a clear winner.  See Table
\ref{ltr}.

\begin{table}
\caption{Letter Recognition, predict letter}
\begin{center}
\begin{tabular}{|l|r|}
\hline
setting & PCC \\ \hline 
PR, 1 & 0.7285\\ \hline 
PR, 2 & {\bf 0.9030} \\ \hline 
KF, default & 0.4484 \\ \hline
KF, units 50,50; dropout 0.1,0.1 & 0.5450 \\ \hline
DN, units 5,5,5 & 0.5268 \\ \hline
DN, units 25,25 & 0.7630 \\ \hline
DN, units 50,50 & 0.7825 \\ \hline
DN, units 200,200 & 0.7620 \\ \hline
\end{tabular}
\end{center}
\label{ltr}
\end{table}

\subsection{New York City Taxi Data}

This is a Kaggle dataset
(https://www.kaggle.com/c/nyc-taxi-trip-duration), in which we predict
trip time.  Results are shown in Table \ref{taxitable}.  There was
perhaps a slight edge to PR over NNs.

\begin{table}
\caption{NYC Taxi, predict trip time}
\begin{center}
\begin{tabular}{|l|r|}
\hline
setting & MAPE \\ \hline 
PR, 1  & {\bf 580.6935}   \\ \hline
PR, 2  & 591.1805  \\ \hline
DN, layers 5,5 & 592.2224   \\ \hline
DN, layers 5,5,5 & 623.5437   \\ \hline
DN, layers 2,2,2 & 592.0192   \\ \hline
\end{tabular}
\end{center}
\label{taxitable}
\end{table}


\subsection{Forest Cover Data}

In this remote sensing study, the goal is to predict the type of ground
cover, among seven classes, from 54 features. There are 581,012
observations, and always guessing the mode (Class 2) would yield 49\%
accuracy. Table \ref{cvrtable} shows PRs and NNs both get about 71\%
right. 
                                                                                
We we unable to run the full degree-2 polynomial, illustrating the
important limitation of PR in {\bf polyreg} mentioned earlier; for
degree 2, our software could not accommodate the size of the polynomial
matrix generated. Section \ref{limits} outlines future work to remedy
this problem.

\begin{table}
\caption{Forest Cover, predict grnd. cover type}
\begin{center}
\begin{tabular}{|l|r|}
\hline
setting & PCC \\ \hline 
PR, 1  & 0.69   \\ \hline
PR, 3  & \textbf{0.80}   \\ \hline
PR, PCA 1 & 0.65 \\ \hline
PR, PCA 2 & 0.69 \\ \hline
PR, PCA 3 & 0.71 \\ \hline
PR, PCA 4,3 & 0.71 \\ \hline
KF, layers 5,5  & 0.72  \\ \hline
NN (reader report) & 0.75 \\ 
\hline
\end{tabular}
\end{center}
\label{cvrtable}
\end{table}

\subsection{MOOCs Data}

This dataset on Harvard/MIT MOOCs was obtained from the
Harvard Dataverse Network, http://thedata.harvard.edu.  Here 
$n = 641138,~ p = 20$.

We wished to predict whether a student will complete the course and
receive certification.  A major challenge of this dataset, though, is
the large number of missing values.  For instance, 58132 of the records
have no value for the {\bf gender} variable.  The only fully intact
variables were {\bf certified}, {\bf nforum\_posts} and the course and
user ID columns.

For the purpose of this paper, we simply used the intact data, adding
four new variables:  The first was {\bf ncert.c}, the total number of
certifications for the given course.  If student A is taking course B,
and the latter has many certifications, we might predict A to complete
the course.  Similarly, we added {\bf ncert.u}, the number of
certifications by the given user, and variables for mean number of forum
posts by user and course.  Altogether, we predicted {\bf certified} from 
{\bf nforum\_posts} and the four added variables.

The results are shown in Table \ref{moocs}.  Note that only about 2.7\%
of the course enrollments ended up certified, so one hopes for an
accuracy level substantially above 0.973.  PR did achieve this, but NNs
did not do so.

\begin{table}
\caption{MOOCs, predict certification}
\begin{center}
\begin{tabular}{|l|r|}
\hline
setting & PCC \\ \hline 
PR, 1  & 0.9871   \\ \hline
PR, 2  & {\bf 0.9870}   \\ \hline
KF, layers 5,5  & 0.9747  \\ \hline
KF, layers 2,2  & 0.9730  \\ \hline
KF, layers 8,8; dropout 0.1 & 0.9712  \\ 
\hline
\end{tabular}
\end{center}
\label{moocs}
\end{table}

\subsection{Crossfit Data}
\label{crossfit}

In this section and the next, we present more detailed analyses.

The focus is on publicly available data from recent Crossfit annual
opens, amateur athletics competitions consisting of five workouts each
year. For each, we fit a neural net and polynomial linear (but not
additive) models. To foreshadow, our PR package, {\bf polyreg}, fit with
a first or second degree polynomial and second degree interactions,
outperforms the NNs slightly in terms of median MAE. (Though the third
degree model did poorly and the fourth degree produced wild estimates
symptomatic of severe collinearity.)



Using {\bf kerasformula}, we built a dense neural network with two
five-node layers (other than the outcome) with ReLU activation. Kernel,
bias, and activity L1-L2 regularization were employed and dropout rate
was set to 40\% and 30\%, respectively. ADAM minimized MSE with batches
of 32. Four separate models were fit, corresponding to polynomial
degrees 1, 2, 3, and 4, using our PR package. All four models, fit by
ordinary least squares, contain two-way interactions. 

For each of four datasets (Men's 2017, Men's 2018, Women's 2017, Women's
2018) we fit 10 models, representing all distinct pairs of opens that
could be used as features to predict each competitor's final rank.
``Rx'', as in ``prescription'', denotes the heaviest weights and most
complex movements. In this analysis, we restrict the population to those
who competed in at least one round at the ``Rx'' level and who reported
their age, height, and weight. The final sample sizes were 118,064
(Men's 2018), 41,103 (Men's 2017), 53,958 (Women's 2018), and 13,815
(Women's 2017). The outcome is rank in the overall competition; like all
other variables, it is scaled 0 to 1.


The results (Table \ref{petetable}) suggest that a first or second
degree polynomial (with two-way interactions) is best in this case in
terms of median mean absolute error (low bias). The first degree model
is preferable because it has lower variance. The third and fourth degree
models are not admissible.

\begin{table}
\caption{Crossfit Open, predict Rx rank}
\begin{center}
\begin{tabular}{|l|r|r|}
\hline
model & MAPE & range among 5 runs \\ \hline 
KF & 0.081 & 0.164  \\ \hline
PR, 1 & 0.070 & \textbf{0.027}  \\ \hline
PR, 2 & 0.071 & 0.069  \\ \hline
PR, 3 & 0.299 & 7.08  \\ \hline
PR, 4 & 87.253 & 3994.5  \\ 
\hline
\end{tabular}
\end{center}
\label{petetable}
\end{table}

\begin{figure}[h]
\centering
\includegraphics[scale=.5]{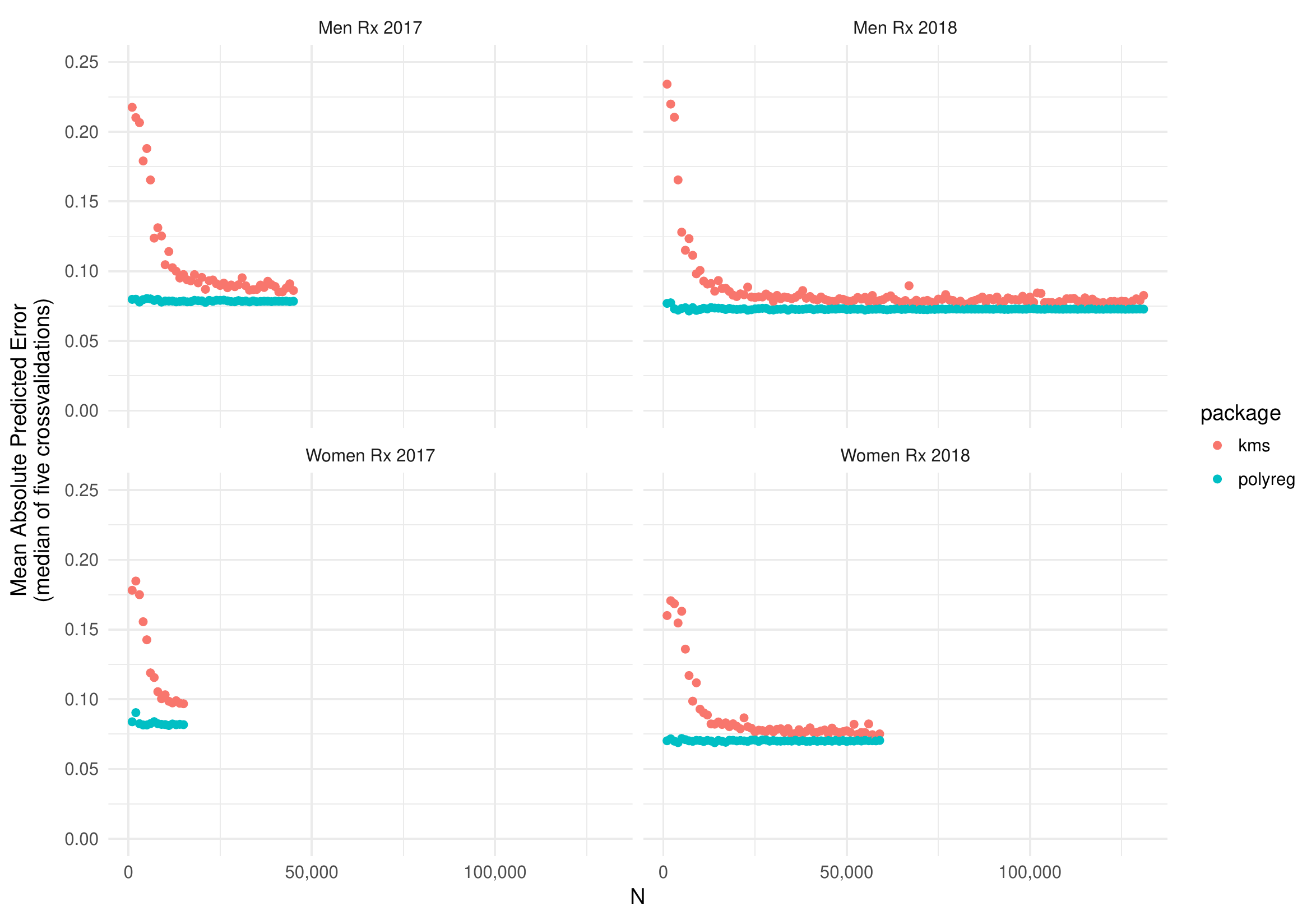}
\caption{Predictive Accuracy by Sample Size}
\label{fig:xfit}
\end{figure}

Next, to assess sample size requirements we compare KF to the best fitting
PR. For each competition, we take subsamples of $1000,\,2000,\,...,
\,N_{open}$ (using only the first two competitions as features).  Figure
\ref{fig:xfit} reports median out-of-sample measures of fit. PR is uniformly
lower though KF nearly converges at fairly modest sample sizes. Notably,
PR's performance is all but invariant--PR performs as well
$N_{subsample}=1,000$ as it does on the full sample for all four
competitions.

\subsection{Big Data and Small Data: New                
Case-Studies in Cancer Genomics}

We construct two cancer datasets from the NCI Genomic Data Commons (GDC)
\cite{GDC}. The first is a compendium of all known cases for the
aggressive brain cancer glioblastoma multiforme (GBM), as well as lower
grade glioma (LGG), with $n$ = 129,119.  The second is of kidney cancer
cases, specifically of papillary renal cell carcinoma ($n$ = 32,457). We
build models that classify patients as `alive' or `dead' based on
genetic mutation type and three variables that assess the impact of the
cancer (as well as patient gender).
The larger sample size notwithstanding, the brain cancer dataset is
considerably more challenging since it does not contain any quantitative
variables.  By contrast, in addition to the impact variables, the kidney
dataset contains patient age at time of diagnosis. The kidney data also
includes patient ethnicity.  The brain cancer data are also higher
entropy (49.94\% of patients are alive compared with 83.68\% in the
kidney cancer data).  As added challenge, the working memory limitations
discussed in the Section \ref{limits} affected this analysis.  

For each data set, we fit six polynomial models, which differed as to
whether second-order interactions were included (in addition to first
order). Models which took advantage of principal components included
quadratic terms; models fit on the mostly qualitative raw data did not.
We fit an NN with {\bf deepnet} with as many hidden nodes as columns in
the model matrix (64 for the brain data and 40 for the kidney). We also
fit an NN with {\bf nnet} of size 10. For each design, we
cross-validated, holding 20\% out for testing (and we report the median
of trials). 	

The results are encouraging for PR (Table \ref{petetable2}). On the
brain cancer data, where polynomial regression might be expected to
struggle, {\bf polyreg} performs as well out of sample as the NNs. On the
kidney cancer data, {\bf polyreg} performs noticeably better than either
{\bf deepnet} or {\bf nnet}.

\begin{table}
\caption{Cancer, predict vital status}
\begin{center}
\begin{tabular}{|l|c|c|}
\hline
model & PCC, brain cancer & PCC, kidney cancer \\ \hline 
deepnet & 0.6587 &0.5387 \\\hline
nnet & {\bf 0.6592} & 0.7170 \\ \hline
PR (1, 1)  &0.6525 &  {\bf 0.8288} \\ \hline
PR (1, 2) & 0.6558 & 0.8265\\ \hline
PR (PCA, 1, 1) & 0.6553 &  0.8271 \\ \hline
PR (PCA, 2, 1) &  0.5336 & 0.7589\\ \hline
PR (PCA, 1, 2) &0.6558 & 0.8270\\ \hline
PR (PCA, 2, 2) & 0.5391 & 0.7840 \\ 
\hline
\end{tabular}
\end{center}
\label{petetable2}
\end{table}

\subsection{MNIST Data}
\label{mnist}

As explained in Section \ref{specialized}, image classification is a
work in progress for us.  We have not yet done much
regarding NN applications in image processing, one of the most
celebrated successes of the NN method.  We will present an example here,
but must stress that it is just preliminary.  In particular, for our
preprocessing stage we simply used PCA, rather than sophisticated
methods such as blocked image smoothing.  It is also a relatively simple
image set, just black-and-white, with only 10 classes.

At the time we ran these experiments, the best reported accuracy for
this data in the literature was 98.6\%.  Now it is over 99\%, but again,
we see here that PR held its own against NNs, in spite of the handicap
of not using advanced image operations.

Our preliminary results for the Fashion MNIST dataset have been similar,
with PR holding its own against NN, in spite of merely using PCA for the
dimension reduction.

\begin{table}
\caption{MNIST classification accuracy}
\begin{center}
\begin{tabular}{|c|c|c|}
\hline
model & PCC \\ \hline
PR 1 & 0.869 \\ \hline
PR 2 & 0.971 \\ \hline
PR 2, 35 PCs & 0.980 \\ \hline
PR 2, 50 PCs & 0.985 \\ \hline
PR 2, 50 PCs & 0.985 \\ \hline
PR 2, 100 PCs & 0.987 \\ \hline
\end{tabular}
\end{center}
\label{mnisttbl}
\end{table}

\subsection{United States 2016 Presidential Election}

We follow the procedure outlined in \cite{mohanty_shaffer}, which uses county-level election results to assess several methods and models, including random forests, causal forests, KRLS, lasso, ridge regression, and elastic net (the last three with and without interactions). The dependent variable is change in GOP vote share in the presidential election, 2012-2016, measured in percentage points (i.e., for each of 3,106 counties, how did Trump's vote share compare to Romney's?). Standard demographics are used as features, as well as which of the nine census regions the county falls in. For each estimator, 100 cross-validations were performed with 80\% of the data used for training and 20\% for testing. Of those estimators, Random Forests have the lowest RMPSE (root mean predicted squared error) at 2.381, though the 95\% confidence interval overlaps slightly with KRLS'. Penalized regression without interactions does the worst, with LASSO coming last at 2.930. 

We add to this experiment with {\bf FSR} (with a minimum of 200 models fit per cross validation), {\bf PR}, and {\bf KF}. For {\bf KF}, the {\bf NN} contains 5 units (but is otherwise at {\bf kms} defaults). Of these, {\bf polyfit} does the best (similar to LASSO), perhaps reflecting the fact that {\bf FSR} sacrifices 20\% of the training data to validation. {\bf KF} comes in last. All three however lagged considerably behind random forests and KRLS (Table \ref{election}).   

\begin{table}
\caption{2016 Presidential Election Results (100 Cross Validations)}
\begin{center}
\begin{tabular}{|c|c|c|}
\hline
 & RMPSE (Mean) & 95\% CI (RMPSE) \\ \hline
\hline
Random Forest & 2.381 &  (2.363, 2.399) \\ \hline
KRLS & 2.415 &  (2.397, 2.432) \\ \hline
LASSO & 2.930 &  (2.916, 2.943) \\ \hline
polyFit & 3.082 & (3.063, 3.100)  \\ \hline
FSR & 3.200 &  (3.173, 3.227) \\ \hline
KF & 3.595 & (3.527, 3.662) \\  \hline
\end{tabular}
\end{center}
\label{election}
\end{table}

\section{Discussion}

The NN $\leftrightarrow$ PR Principle, combined with our experimental
results, gives rise to a number of issues, to be discussed in this
section.

\subsection{Effects of Regularization}

Many practitioners tend to initialize their networks to large numbers of
units, and possibly large numbers of layers, reflecting a recommendation
that large networks are better able to capture non-linearities
\cite{friedman2001elements}.  However, that implies a need for
regularization, which {\bf kerasformula} does for kernel, bias, and
activity terms by default. 

One of the reasons for the popularity of $L_1$ regularization is that it
tends to set most weights to 0, a form of dimension reduction.  But it
has a particularly interesting effect in NNs:

If most weights are 0, this means most of the units/neurons in the
initial model are eliminated.  In fact, entire layers might be
eliminated.  So, a network that the user specifies as consisting of,
say, 5 layers with 100 units per layer may in the end have many fewer
than 100 units in some layers, and may have fewer than 5 layers.

Some preliminary investigations conducted by the authors showed that
many weights are indeed set to 0.  See Table \ref{zeros}.  Note in
particular that the larger the initial number of units set by the user,
the higher the proportion of 0 weights.  This is consistent with our
belief that many users overfit.  And note carefully that this is not
fully solved by the use of regularization, as the initial overfitting is
placing one additional burden on the estimative ability of the NN (or
other) system.

\begin{table}
\caption{Zeros among MNIST weights}
\begin{center}
\begin{tabular}{|l|r|r|}
\hline
layer & neurons & prop.\ 0s \\ \hline
\hline
1 & 256 & 0.8072 \\ \hline
2 & 128 & 0.6724 \\ \hline
3 & 64 & 0.5105 \\ \hline
\end{tabular}
\end{center}
\label{zeros}
\end{table}

\subsection{Extrapolation Issues}

One comment we have received in response to our work here is that
polynomials can take on very large values at the edges of a dataset, and
thus there is concern that use of PR may be problematic for predicting
new cases near or beyond the edges.

This of course is always an issue, no matter what learning method one
uses.  But once again, we must appeal to the NN $\leftrightarrow$ PR
Principle: Any problem with PR has a counterpart in NNs.  In this case,
this means that, since the NN fit at each layer will be close to a
polynomial, NNs will exhibit the same troublesome extrapolation behavior
as PR.  Indeed, since the degree of the polynomial increases rapidly
with each layer, extrapolation problems may increase from layer to
layer.

On the other hand, note too that, regardless of method used,
extrapolation tends to be less of an issue in classification
applications.  Consider a two-class setting, for example.  In typical
applications, the conditional probability of class 1 will be close to 0
or 1 near the edges of the predictor space, so large values of the
estimated $r(t)$ won't be a problem; those large values will
likely become even closer to 0 or 1, so that the actual predicted value
will not change.

A similar way to view this would be to consider support vector
machines.  The kernel function is either an exact or approximate
polynomial, yet SVMs work well.

\subsection{NNs and Overfitting} 

It is well-known that NNs are prone to overfitting
\cite{chollet2018deep}, which has been the subject of much study, e.g.\
\cite{giles}.  In this section, we explore the causes of this problem,
especially in the context of the NN $\leftrightarrow$ PR Principle.
These considerations may explain why in some of the experiments reported
here, PR actually outperformed NNs, rather than just matching them.  

In part, overfitting stems from the multitude of hyperparameters in
typical implementations.  As one tries to minimize the objective
function, the larger the number of hyperparameters, the more likely
that the minimizing configuration will seize upon anomalies in the
training set, hence overfitting.

One popular technique to counter overfitting in neural networks is
\textit{dropout} \cite{srivastava}.  This ``thins out'' the network by
randomly culling a certain proportion of the neurons.  Note, though,
that if the dropout rate is to be determined from the data, this is yet
another hyperparameter, compounding the problem.

%

%
%

\subsection{Limitations and Remedies}
\label{limits}

As mentioned, the PR method, while effective, has potential limitations
in terms of memory, run time and multicollinearity.  In this section, we
discuss existing and future remedies.

To set the stage, let's take a closer look at the problems.  As before,
let $n$ and $p$ denote the number of cases and the number of
predictors/features, respectively.  Denote the degree of the polynomial
by $d$.

First, how large will the polynomial matrix be?  It will have $n$ rows;
how many columns, $l_d$, will there be?  This can be calculated exactly,
but a rough upper bound will be handier.  With $d = 1$, the number of
possible terms $l_d$ is about $p$.  What happens when we go to degree
$d+1$ from degree $d$?  Consider one of the $p$ variables $x_i$.  We can
form $l_d$ new terms by multiplying each of the existing ones by $x_i$.
So, we have $l_{d+1} \approx (p+1) l_d$.  That implies that $l_d$ is
$O(p^d)$.  So the polynomial matrix can be large indeed.

The computation for the linear and generalized linear model (e.g.\
logistic) involves inversion (or equivalent) of an $l_d \times l_d$
matrix.  Using QR decomposition, this takes $O(n,l_d^2)$
time.\footnote{This notation is a bit nonstandard.  We do not use a
product notation, as the timing is not multiplicative.  The time depends
on $n$ and $l_d \times l_d$, but separately.} In the classification
case, this becomes $O(n,q l_d^2)$, where $q$ is the number of classes.

Beyond that, there is the statistical issue.  The results of
\cite{portnoy} would suggest that we should have $l_d < \sqrt{n}$.
Though modern research results on the LASSO and the like are more
optimistic, it is clear that we need to keep $d$ small unless $n$ is
extremely large.  This is consistent with our empirical findings here,
in which we found that $d = 2$ is sufficient in most cases.

Dimension reduction via PCA remedies time, space, and multicollinearity
problems and was found to work well in the case of Million Song dataset.
Another possible form of dimension reduction would be to mimic the
dropout ``lever'' in NNs.  Another possibility would be to apply PCA to
the matrix of polynomial terms, rather than to the original data as is
presently the case; this may help with nonmonotonic data.  Other
approaches to nonmonotonicity may be nonlinear PCA \cite{zhang} and
nonnegative matrix factorization \cite{lee}.

One can also delete randomly-selected columns of the polynomial features
matrix. Our {\bf polyreg} package does have such an option.

Multicollinearity itself might be handled by ridge regression or the LASSO.
It should be kept in mind, though, that our results indicate the
multicollinearity is manifested in NNs as well, and is a possible sign of
overfitting in both.  See related work in \cite{shin}.

Parallel computation is definitely a possibility.  The software
currently provides the option of parallelizing PR in the classification
case.  Both time and space issues might be resolved by using the
Software Alchemy technique of \cite{nmsa}.  Physical memory limitations
could be resolved using the backing store feature of the {\bf bigmemory}
package.

\section{Conclusions and Future Work}

We have presented a new way to look at NNs, as essentially a form of PR.
Though some previous work had noted some other kinds of connection of
NNs to polynomials, we presented a simple analytic argument involving
the activation function showing the connection explicitly and in a very
strong manner.

We have shown that viewing NNs as PR models reveals properties of NNs
that are, to our knowledge, new to the field, and which should be useful
to practitioners.  For instance, our NN $\leftrightarrow$ PR Principle
predicted that NNs should have multicollinearity problems in later
layers, which in turn suggests ways to avoid convergence problems.
Among other things, this suggests some useful improvements in diagnostic
features in NN software.

Most importantly, we have shown that PR should be an effective {\it
alternative} to NNs.  We performed experiments with a variety of data of
various types, and in {\it all} cases, PR performed either similarly to,
or substantially better than, NNs --- without the NN troubles of trying
to find good combinations of hyperparameters.  

The fact that in some cases PR actually outperformed NNs reflects the
fact that NNs are often overparameterized, in essence fitting a
higher-degree polynomial than they should.

Much remains to be done.  The problems and remedies outlined in the
preceding section need to be tested and implemented; the phenomenon of
multicollinearity in NNs needs thorough investigation; more
experimentation with large-$p$ datasets should be conducted; and the
approach here needs to be integrated with preprocessing of images, text
and so on as with, e.g., CNNs.

It is conceivable that PR may be competitive with other machine learning
techniques, such as random forests, SVM and so on.  We have focused on
NNs here because of the direct connection to PR described in Section
\ref{why}, but similar connections to other methods could be explored.

\bibliographystyle{acm}
\bibliography{arXiv.bib}   

\end{document}